%% file: paper.tex
\documentclass[10pt,twocolumn,letterpaper]{article}

\usepackage{cvpr}
\usepackage{times}
\usepackage{epsfig}
\usepackage{graphicx}
\usepackage{amsmath}
\usepackage{amssymb}

\usepackage{float}

\usepackage[ruled,linesnumbered]{algorithm2e}

\usepackage{amsthm}
\newtheorem{mydef}{Definition}[section]

\usepackage{subcaption}

\usepackage[pagebackref=true,breaklinks=true,letterpaper=true,colorlinks,bookmarks=false]{hyperref}

\cvprfinalcopy 

\def\cvprPaperID{13} 

\ifcvprfinal\fi
\begin{document}

\title{AnonymousNet: Natural Face De-Identification with Measurable Privacy}

\author{Tao Li\\
Department of Computer Science\\
Purdue University\\
{\tt\small taoli@purdue.edu}
\and
Lei Lin\\
Goergen Institute for Data Science\\
University of Rochester\\
{\tt\small lei.lin@rochester.edu}
}

\maketitle

\input{sec_abstract}

\input{sec_introduction}

\input{sec_literature}

\input{sec_definition}

\input{sec_method}

\input{sec_experiment}

\input{sec_conclusion}

\input{sec_acknowledgment}

{\small
\bibliographystyle{ieee}
\bibliography{db}
}

\end{document}

%% file: sec_abstract.tex
\begin{abstract}
With billions of personal images being generated from social media and cameras of all sorts on a daily basis, security and privacy are unprecedentedly challenged.
Although extensive attempts have been made, existing face image de-identification techniques are either insufficient in photo-reality or incapable of balancing privacy and usability qualitatively and quantitatively, i.e., they fail to answer counterfactual questions such as ``is it private now?'', ``how private is it?'', and ``can it be more private?''
In this paper, we propose a novel framework called AnonymousNet, with an effort to address these issues systematically, balance usability, and enhance privacy in a natural and measurable manner.
The framework encompasses four stages:
facial attribute estimation,
privacy-metric-oriented face obfuscation,
directed natural image synthesis,
and adversarial perturbation.
Not only do we achieve the state-of-the-arts in terms of image quality and attribute prediction accuracy,
we are also the first to show that facial privacy is measurable, can be factorized, and accordingly be manipulated in a photo-realistic fashion to fulfill different requirements and application scenarios.
Experiments further demonstrate the effectiveness of the proposed framework.
\end{abstract}

%% file: sec_introduction.tex
\section{Introduction}
The deployment of internet of things devices, such as surveillance cameras and sensors, are rising dramatically in recent years.
Particularly, the popularity of smartphones allows billions of photos being uploaded to social networks and shared among people on a daily basis. Although the blooming development has advanced machine learning applications to bring convenience and enhance user experience, it may capture confidential information incidentally and increases the risks of privacy leakage.
To protect privacy, the most straightforward approach is access control \cite{sandhu1994access}, such as the ``Restrict Others'' setting in Facebook \cite{li2017blur}.
Cryptography techniques such as encryption and secure multi-party computation \cite{li2019autompc} can also be applied to mitigate privacy threats.
In computer vision community, privacy-enhancing technologies are mainly obfuscation-based; for example, obfuscating sensitive information like faces and numbers in an image by using traditional approaches including blurring, pixelation, and masking (see Figure~\ref{fig:compare}).
However, there are at least two drawbacks with these traditional approaches. First, researchers have  shown these techniques are vulnerable. The faceless person recognition system proposed in \cite{oh2016faceless} can be trained with limited training samples and then identify the target in an obfuscated image by using body features. Another study also shows that deep learning models can successfully identify faces in images encrypted with these techniques with high accuracies \cite{mcpherson2016defeating}.  Second, images processed with these techniques result in unsatisfying perception in general. A study shows that both blurring and blocking will impact image perception scores, and even lower scores are observed for images obfuscated by blocking \cite{li2017blur}.

\input{fig_intro}
\input{fig_compare}
On the other hand, new techniques and mechanisms are being applied to enhance image obfuscation. A game theory framework called Adversarial Image Perturbation has been proposed to determine effective image obfuscation approach when the choice of countermeasures is unknown \cite{oh2017adversarial}.
Recently, the generative adversarial networks (GAN) can generate realistic natural images following original input data distribution via adversarial training \cite{goodfellow2014generative}, therefore it has become more and more popular for novel image obfuscation techniques. \textit{Wu et al.} \cite{wu2018privacy} developed a GAN with two newly designed modules verificator and regulator for face de-identification. Considering subjects in social media photos appear in diverse activities and head orientations, \textit{Sun et al.} \cite{sun2018natural} proposed a two-stage model to inpaint the head region conditioned on facial landmarks.

Note that there exists a tradeoff between privacy protection and dataset usability \cite{zhang2018privacy,hasan2018viewer}. High obfuscation levels fail to preserve utility for various tasks while low obfuscation levels lead to recognition of private information.
Unfortunately, current methods are unable to find a way to quantify this matter; neither can they be adapted with correspondence to various privacy metrics nor real-world scenarios under different requirement settings.
To tackle these, we propose the AnonymousNet, a four-stage frameworks consisted of facial semantic extraction powered by a deep Convolutional Neural Network \cite{krizhevsky2012imagenet};
a attribute selection method with regards to privacy metrics such as $k$-anonymity \cite{sweeney2002achieving}, $l$-diverse \cite{machanavajjhala2006diversity}, and $t$-closeness \cite{li2007t};
a Generative Neural Network \cite{goodfellow2014generative} for photo-realistic image generator; and a universal adversarial perturbation \cite{moosavi2017universal} to mitigate potential security and privacy threats.


The rest of the paper is organized as follows: Section \ref{sec:literature} provides necessary background in privacy-preserving data mining and reviews recent advances in facial image editing;
Section \ref{sec:definition} formalizes the face de-identification problem and introduce privacy metrics;
Section \ref{sec:method} outlines the four-stage AnonymousNet framework, including facial feature extraction, semantic-based attribute obfuscation, de-identified face generation, and adversarial perturbation;
Section \ref{sec:experiment} details experiment settings and evaluate the results;
we conclude this paper in Section \ref{sec:conclusion} by discussions of future research directions.

%% file: fig_intro.tex
\begin{figure}[t]
\begin{center}
    \begin{subfigure}[b]{0.11\textwidth}
        \includegraphics[width=\linewidth]{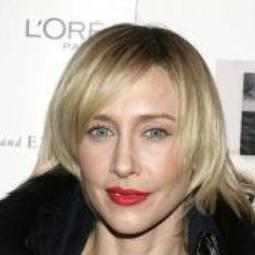}
        \includegraphics[width=\linewidth]{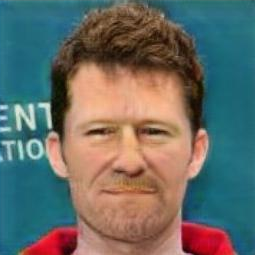}
        \caption{Original}
    \end{subfigure}
    \begin{subfigure}[b]{0.11\textwidth}
        \includegraphics[width=\linewidth]{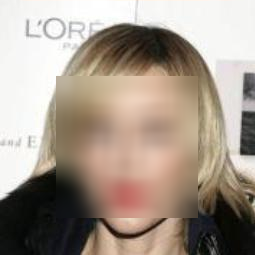}
        \includegraphics[width=\linewidth]{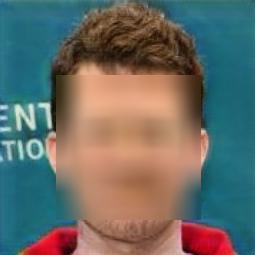}
        \caption{Blurring}
    \end{subfigure}
    \begin{subfigure}[b]{0.11\textwidth}
        \includegraphics[width=\linewidth]{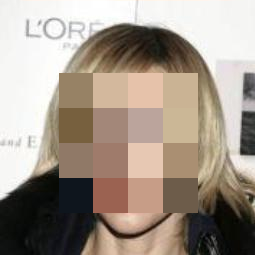}
        \includegraphics[width=\linewidth]{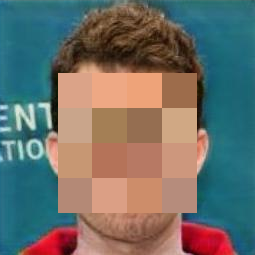}
        \caption{Pixelation}
    \end{subfigure}
    \begin{subfigure}[b]{0.11\textwidth}
        \includegraphics[width=\linewidth]{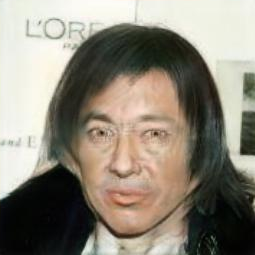}
        \includegraphics[width=\linewidth]{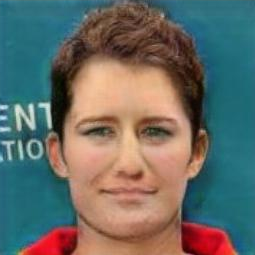}
        \caption{Ours}
    \end{subfigure}
\end{center}
\caption{A brief comparison of obfuscation methods. Our approach not only de-identifies an image by synthesizing a photo-realistic alternative, but also provides a controllable and measurable way for privacy preservation. Moreover, an adversarial perturbation is introduced to further enhance security and privacy against malicious detectors.}
\label{fig:intro}
\end{figure}

%% file: fig_compare.tex
\begin{figure*}[t]
\begin{center}
    \begin{subfigure}[b]{0.12\textwidth}
        \includegraphics[width=\linewidth]{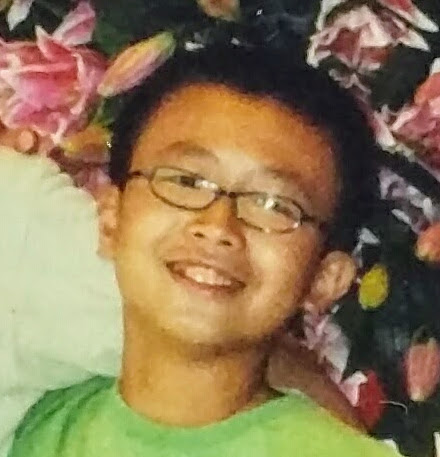}
        \includegraphics[width=\linewidth]{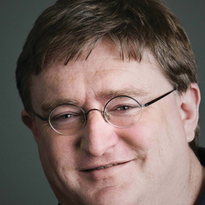}
        \includegraphics[width=\linewidth]{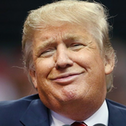}
        \caption{Original}
        \label{fig:original}
    \end{subfigure}
    \begin{subfigure}[b]{0.12\textwidth}
        \includegraphics[width=\linewidth]{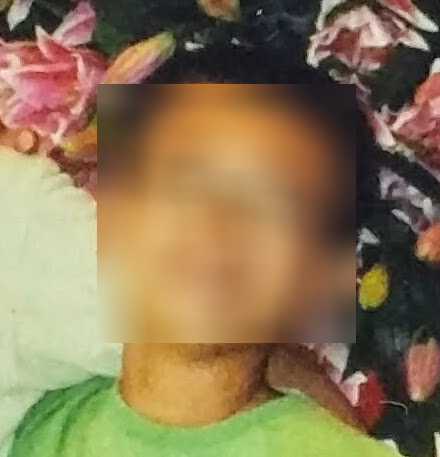}
        \includegraphics[width=\linewidth]{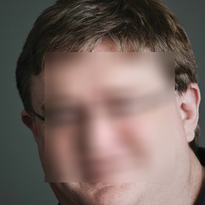}
        \includegraphics[width=\linewidth]{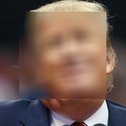}
        \caption{Blurring}
        \label{fig:blurring}
    \end{subfigure}
    \begin{subfigure}[b]{0.12\textwidth}
        \includegraphics[width=\linewidth]{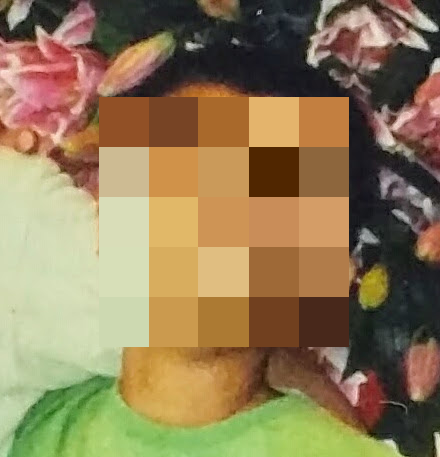}
        \includegraphics[width=\linewidth]{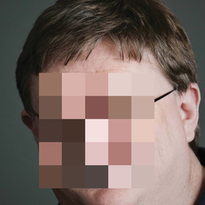}
        \includegraphics[width=\linewidth]{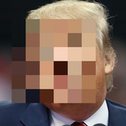}
        \caption{Pixelation}
        \label{fig:pixelation}
    \end{subfigure}
    \begin{subfigure}[b]{0.12\textwidth}
        \includegraphics[width=\linewidth]{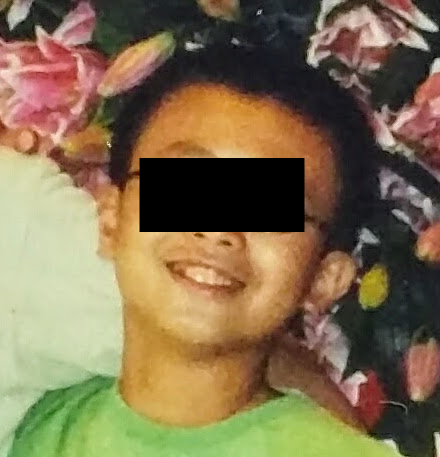}
        \includegraphics[width=\linewidth]{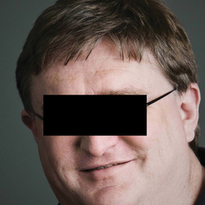}
        \includegraphics[width=\linewidth]{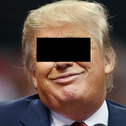}
        \caption{Masking}
        \label{fig:masking}
    \end{subfigure}
    \begin{subfigure}[b]{0.12\textwidth}
        \includegraphics[width=\linewidth]{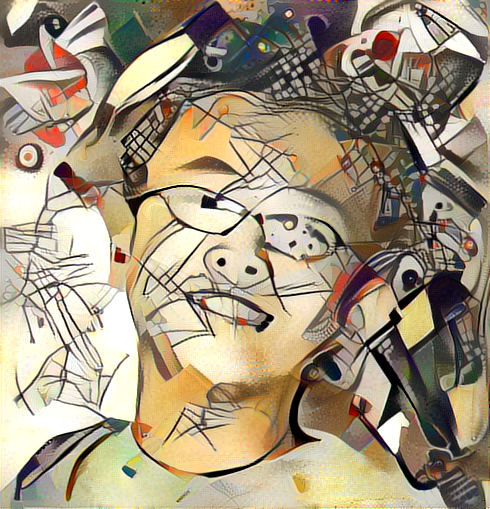}
        \includegraphics[width=\linewidth]{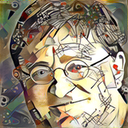}
        \includegraphics[width=\linewidth]{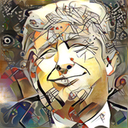}
        \caption{Abstract}
        \label{fig:abstract}
    \end{subfigure}
    \begin{subfigure}[b]{0.12\textwidth}
        \includegraphics[width=\linewidth]{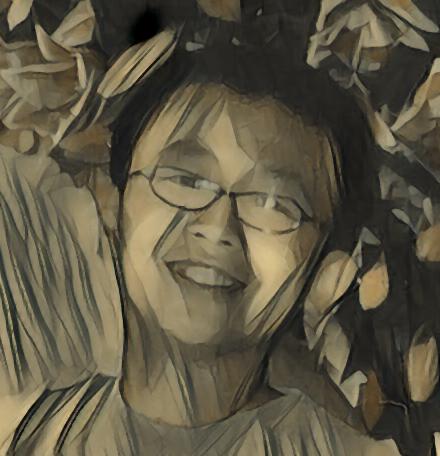}
        \includegraphics[width=\linewidth]{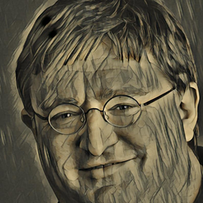}
        \includegraphics[width=\linewidth]{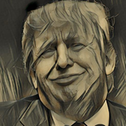}
        \caption{Portrait}
        \label{fig:portrait}
    \end{subfigure}
    \begin{subfigure}[b]{0.12\textwidth}
        \includegraphics[width=\linewidth]{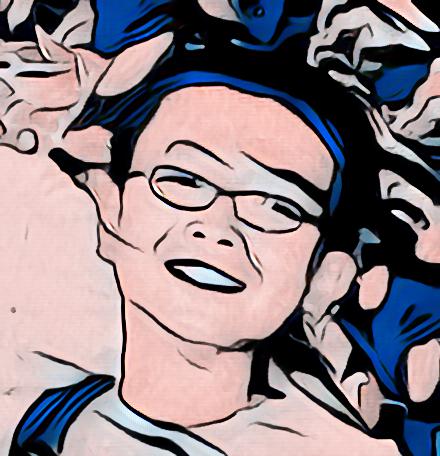}
        \includegraphics[width=\linewidth]{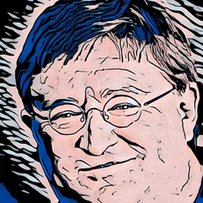}
        \includegraphics[width=\linewidth]{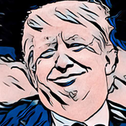}
        \caption{Cartoon}
        \label{fig:cartoon}
    \end{subfigure}
    \begin{subfigure}[b]{0.12\textwidth}
        \includegraphics[width=\linewidth]{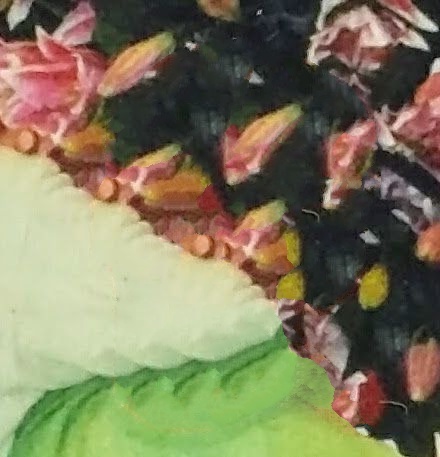}
        \includegraphics[width=\linewidth]{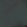}
        \includegraphics[width=\linewidth]{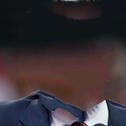}
        \caption{Inpainting}
        \label{fig:inpainting}
    \end{subfigure}
\end{center}
\caption{Comparison of canonical image processing methods for face obfuscation. From left to right are blurring, pixelation, masking, deep Convolutional Neural Network-based style transfer \cite{johnson2016perceptual} (abstract painting style, portrait painting style, and cartoon style \cite{hasan2017cartooning}), and Generative Adversarial Network-based image inpainting \cite{yeh2017semantic}.}
\label{fig:compare}
\end{figure*}

%% file: sec_literature.tex
\section{Related Work}\label{sec:literature}

\paragraph{Facial Landmark Detection}
Photo privacy protection includes two important aspects: sensitive content detection and obfuscation method \cite{li2018human}. Facial information is one of the highest sensitive contents. Various head poses, orientations, lighting conditions and so on make photo privacy protection more challenging.
Facial landmark detection has been an active research field because it is the first important step for other facial-related applications. In the past few decades, numerous algorithms like support vector machine \cite{rapp2011multiple} and random forecast \cite{amberg2011optimal} have been proposed. Despite these models have made significant progress, they rely on handcrafted local features and may not operate under a wide range of conditions like illumination, occultation, poses.

Recently deep learning models have made huge advances to deal with issues like occultation and variability in facial landmark detection.
\cite{sun2013deep} proposed a cascaded convolutional network for facial keypoint detection which can utilize the texture context information over the entire face and encode geometric constraints implicitly. Further, a deep multi-task learning framework is presented that combine facial landmark detection with correlated tasks like head pose estimation \cite{zhang2014facial}. \textit{Dong et al.} (2018) adopted a generative adversarial network to transform facial images into style-aggregated ones, which are then deployed together to train a facial landmark detector \cite{dong2018style}.

\paragraph{Image Inpainting}
Previous study shows that comparing with image blurring and blocking, image inpainting provides a more effective and better user experience \cite{li2018human}. Face replacement has become a popular image inpainting technology for privacy protection. \cite{bitouk2008face} created a large face library from public internet. Given an input face, the most similar one in the library will be chosen for face replacement. \cite{min2010automatic} applied a 2D morphable model to adjust the shape of a source face to match the target face. \cite{lin2012face} proposed a framework to generate a personalized 3D head model from one frontal face. The 3D head model can then be rendered at any pose to swap with the face in the target image.
Figure~\ref{fig:compare} compares several classic image processing methods for face identity obfuscation.

\paragraph{GAN-based Face Generation}
Generative Adversarial Network (GAN) is a system of two neural networks who contests with each other under a zero-sum game setting.
GAN was first introduced by \textit{Goodfellow et al.} \cite{goodfellow2014generative} in 2014. Since then, great progresses have been made, as shown in Figure \ref{fig:GAN}:
in 2015, \textit{Radford et al.} \cite{radford2015unsupervised} designed deep convolutional generative
adversarial networks (DCGANs) which demonstrated the adversarial pair (both the generator and discriminator) can learn a hierarchy of representations from object parts to scenes;
\textit{Liu et al.} \cite{liu2016coupled} proposed coupled generative adversarial network (CoGAN) in 2016, which is capable of learning a joint distribution with only samples from marginal distributions and without tuple of corresponding images from different domains;
\textit{Karras et al.} \cite{karras2017progressive} described a new training method for GAN in 2017, whose main idea is to train generator and discriminator progressively by starting from a low resolution and adding new layers of the model with fine details incrementally;
and more recently, \textit{Karras et al.} \cite{karras2018style}  proposed a style-based generator architecture (StyleGAN) that is able to learn high-level facial attributes in a automated and unsupervised manner, generates images with stochastic variations, and achieves the state-of-the-art.
\input{fig_GAN}

\paragraph{Privacy-Preserving Data Mining}
Theories and practices of privacy-preserving techniques have been studied extensively in the database and data mining communities; to name a few: \cite{agrawal2000privacy,clifton2002tools,vaidya2002privacy}.
To measure the disclosure risk of an anonymized table of a database, \textit{Samarati \& Sweeney} (1998) \cite{samarati1998protecting} introduced the $k$-anonymity property such that each record in the database is indistinguishable with at least $k-1$ records. Although being sufficient to protect against identity disclosure, $k$-anonymity is limited to prevent attribute disclosure. In regards to that, \textit{Machanavajjhala et al.} (2006) \cite{machanavajjhala2006diversity} introduced a new privacy property called $l$-diversity, which requires protected attributes to have at least $l$ well-represented values in each equivalence class.
\textit{Newton et al.} (2005) \cite{newton2005preserving} introduced the first privacy-enabling algorithm, $k$-Same, to the context of image databases.
\textit{Gross et al.} (2005) \cite{gross2005integrating} demonstrated a tradeoff between disclosure risk (i.e., image obfuscation level) and classification accuracy. To tackle this, they introduced $k$-Same-Select to balance privacy and usability.
\textit{Zhang} (2018) \cite{zhang2018privacy} further designed a OBFUSCATE function that adds random noises to existing samples or creates new samples, in an effort to hide sensitive information in the dataset while preserving model accuracy.

%% file: fig_GAN.tex
\begin{figure}[h]
\begin{center}
\includegraphics[width=\linewidth]{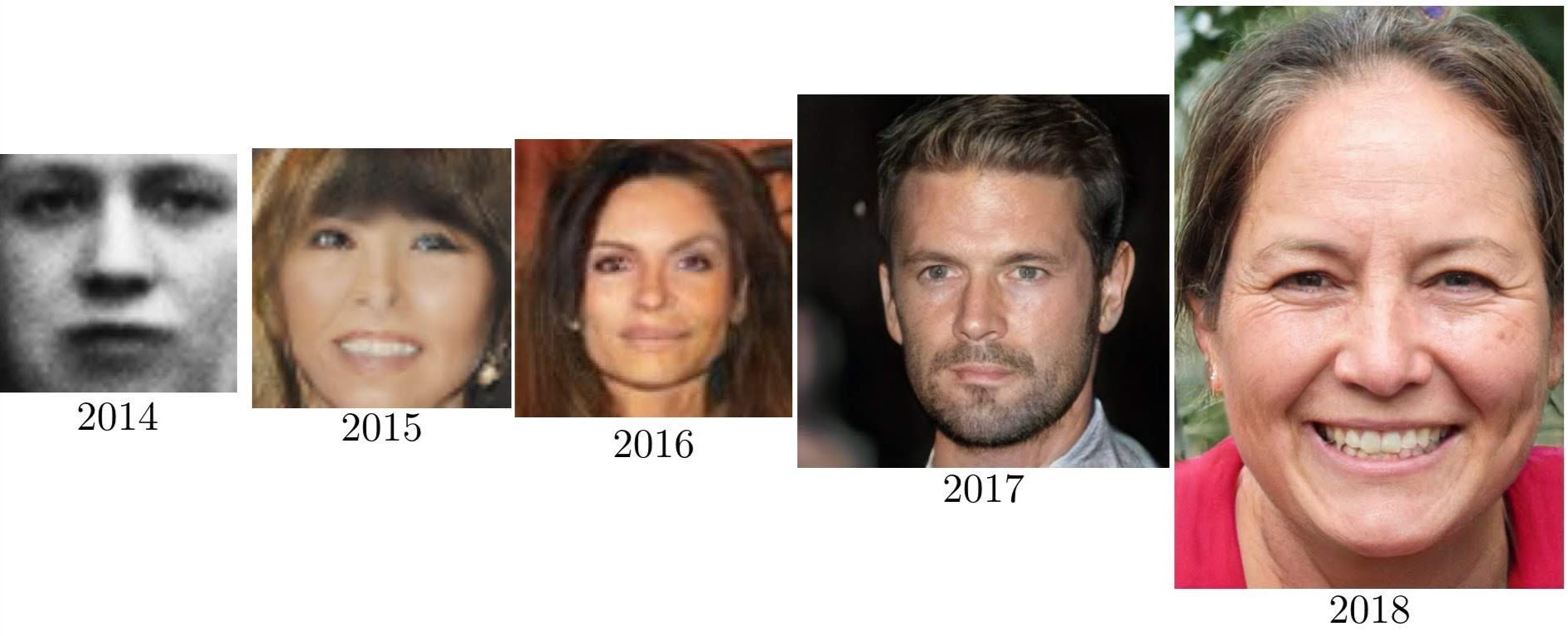}
\end{center}
\caption{Progresses of GAN-based face generation since 2014 \cite{goodfellow2019} (from left to right: GAN (2014) \cite{goodfellow2014generative},
DCGAN (2015) \cite{radford2015unsupervised},
CoGAN (2016) \cite{liu2016coupled},
\textit{Karras et al.} (2017) \cite{karras2017progressive},
and StyleGAN (2018) \cite{karras2018style}).}
\label{fig:GAN}
\end{figure}

%% file: sec_definition.tex
\section{Preliminary}\label{sec:definition}

\subsection{Formulation of Face De-Identification}\label{sec:formulation}
We first formally define the problem of face de-identification,
extending notations and definitions from \textit{Newton et al.} \cite{newton2005preserving}.
This process is important as it helps us to precisely define the methods and to build a solid foundation for following discussions of theoretical properties.
\begin{mydef}
\textbf{(Face Image).}
A face image (or simple `face' or `image') is a 3D matrix $I$ of $m$ columns, $n$ rows, and $c$ channels. $c$ is usually $3$ in common color spaces (e.g., RGB and YUV). Each cell in $I$ stores a color coding for a pixel, ranging from $0$ to $255$ inclusively. A face image contains a normalized image of only one person's face.
\end{mydef}

\begin{mydef}
\textbf{(Face Set).} A face set is a set of $M$ face images:
$\{ \Gamma_i : i = 1, \dots, M \}$.
\end{mydef}

\begin{mydef}
\textbf{(Person-Specific Face Set).} Let $\mathbf{H}$ be a face set of
$M$ face images, $\{ \Gamma_i, \dots, \Gamma_M \}$. $\mathbf{H}$ is said to be person-specific if and only if each $\Gamma_i \in \mathbf{H}$ only relates to
one person and $\Gamma_i \ne \Gamma_j$ for any $i \ne j$.
\end{mydef}

\begin{mydef}
\textbf{(Face De-Identification Function).} Let $\mathbf{H}$ and $\mathbf{H_d}$ be person-specific face set.
\begin{equation}
f: \mathbf{H} \rightarrow \mathbf{H_d}
\end{equation}
is called face de-identification function if it attempts to obfuscate the identity of the original face image.
\end{mydef}

\begin{mydef}
\textbf{(De-Identified Face).}
Given $\Gamma \in \mathbf{H}$ and de-identification function $f$,
$\Gamma_d \in \mathbf{H_d}$ is a $f$ de-identified face of $\Gamma$ if
\begin{equation}
\Gamma_d = f(\Gamma)
\end{equation}
\end{mydef}
Figure \ref{fig:compare} illustrates several canonical image processing methods for face de-identification, including blurring, pixelation, masking, deep Convolutional Neural Network-based style transfer \cite{johnson2016perceptual} (abstract painting style, portrait painting style, and cartoon style \cite{hasan2017cartooning}), and Generative Adversarial Network-based image inpainting \cite{yeh2017semantic}.

\subsection{Privacy Metrics}\label{sec:metric}
The motivation of privacy metrics is to provide qualitative and quantitative measurement of the degree of privacy enjoyed by specific users (personal images in our case) and the proper amount of privacy protection offered with correspondence to trade-offs in usability. This section, we only discuss metrics that are used in our framework. \cite{wagner2018technical} provides a more comprehensive list for interested readers.

\paragraph{$k$-Anonymity}
$k$-anonymity is a widely applied metric to evaluate a dataset's level of anonymity \cite{samarati1998protecting}. It requires that each record in the dataset is indistinguishable with at least $k-1$ other records with respect to quasi-identifiers, which refer to attributes that can potentially be taken together to identify an individual like zip code or birth date \cite{li2007t}. In the case of a face dataset, these quasi-identifiers may include semantic attributes like eyeglasses, pointy nose, and oval face,  and so on. If a dataset satisfies the condition of $k$-Anonymity, with only quasi-identifiers of one individual known, the true record can only be chosen with a probability of $1/k$.

However, there are some scenarios that $k$-anonymity cannot provide enough protection. For example, for $k$ subjects that have the same values for the quasi-identifiers, if they all have the same sensitive attribute like heart disease, an adversary can be certain that the target identify in the $k$ subjects must have heart disease. The $k$-Anonymity thus fails to protect sensitive information from the homogeneity attack \cite{li2007t}. 

\paragraph{$l$-Diversity}
$l$-diversity is proposed to address the limitations of $k$-anonymity \cite{machanavajjhala2006diversity}. The basic concept is that for the equivalence class representing a set of records with the same values for the quasi-identifiers, it should have at least $l$ ``well-represented'' values for the sensitive attribute.

The most straightforward definition of ``well-represented'' values is to ensure the equivalence class has $l$ distinct values for the sensitive attribute. In this definition, the frequencies of $l$ distinct values are not considered. An adversary may conclude that the sensitive attribute of a targeted identity has the value with the highest frequency. Therefore there is a stronger definition of $l$-diversity named Entropy $l$-diversity, which is defined as follows:
\begin{equation}
Entropy(E) \geq \log l
\end{equation}
\begin{equation}
Entropy(E) = -\sum_{s \in S}p(E, s) \log p(E, s)
\end{equation}
where $E$ is the equivalence class, $S$ is the value set of the sensitive attribute, and $p(E, s)$ is the fraction of records in $E$ that have sensitive value $s$.

\paragraph{$t$-Closeness} Adversaries sometimes have knowledge of the global distribution of sensitive attributes, for example, the distributions of facial attributes are easy to obtain (see Figure~\ref{fig:CeleA}). To prevent privacy disclosure by an adversary with such knowledge, \cite{li2007t} introduced $t$-closeness, which updates $k$-anonymity with correspondence to the distribution of sensitive values, requiring that the distribution $S_E$ of sensitive values in any equivalence class $E$ must be close to their distribution $S$ in the entire database, i.e.,
\begin{equation}\label{eqn:t-closeness}
\forall E : d(S, S_E) \le t
\end{equation}
where $d(S, S_E)$ is the distance between distribution $S$ and $S_E$ measured by the Earth Mover Distance \cite{rubner2000earth} and $t$ is the privacy threshold at which $d(S, S_E)$ should not exceed.

\paragraph{Randomness} 
Randomization is another approach to protect data privacy. It is realized by adding random noise to existing samples \cite{zhang2018privacy}. Given an individual sample, we can randomly select part of its features with a ratio of $\gamma$, and add Gaussian noise $N(0, \sigma)$. Instead of random feature selection, we can also identify sensitive features first, and add Gaussian noise $N(0, \sigma)$. Another randomization approach is to introduce new samples into the dataset \cite{zhang2018privacy}. To generate a new sample, we can randomly pick an original sample from the dataset, revise its feature value and add small amount of noise at the same time. For example, the value $x_i$ of pixel $i$ in an original image can be replaced with $255-x_i$ with Gaussian noise $N(0, \sigma)$ added in the new sample.
From a broader perspective, adversarial perturbation can also be consider as a randomization method.

%% file: sec_method.tex
\input{fig_stage1}
\section{The AnonymousNet}\label{sec:method}
In an effort to obfuscate facial identities and generate photo-realistic alternatives, balance privacy and usability qualitatively and quantitatively, answer counterfactual questions such as ``is it private now?'', ``how private is it?'', and ``can it be more/less private?'', and finally achieve controllable and measurable privacy, we propose the AnonymousNet framework, which encompasses four stages: facial feature extraction, semantic-based attribute obfuscation, de-identified face generation, and adversarial perturbation, as detailed below.

\subsection*{Stage-I: Facial Attribute Prediction}\label{sec:stage1}
We adopt GoogLeNet \cite{szegedy2015going} for facial attribute extraction, which consists of $22$ layers witn $9$ Inception blocks and excelled in ImageNet Large-Scale Visual Recognition Challenge 2014 \cite{russakovsky2015imagenet}. Unlike most other classifiers in ImageNet \cite{deng2009imagenet}, GoogLeNet is not trained for one label, but rather is fed with $40$ facial attributes at the same time and outputs multiple classification results accordingly. GoogLeNet further leverages a trick of using $1 \times 1$ kernels to increase depth while reducing dimension and reserving computational resources.
Figure \ref{fig:stage1} outlines the facial attribute prediction pipeline, where labelled images are first fed into the model, features are extracted from the fully connected (FC) layer, and then $40$ random forest classifiers \cite{liaw2002classification} are trained and facial attributes are subsequently obtained.

\subsection*{Stage-II: Privacy-Aware Face Obfuscation}\label{sec:stage2}
Provided with semantic information of each facial image as well as attribute distribution over the entire database (see Figure~\ref{fig:CeleA}), we are one step closer towards our goal: face de-identification with privacy guarantees.
We first consider a toy example: database consist of $2$ identities, both of which share a common attribute \texttt{<Male>}. In this case, modifying the gender attribute will not change the level of privacy, since altering this attribute will not change the possibility of guessing the identity given gender.
Consider another example that for the same database and each entity has three boolean attributes: \texttt{<Male, Big\_Nose, Black\_Hair>} and one identity has black hair and the other does not. For this case, both of them should be updated to either black hair or non-black hair since the identity will be revealed if the hair color is known.
These two example provide us insights of how to select facial attributes such that $k$-anonymity and $l$-diversity are satisfied.

\input{algo_PPAS}
However, as discussed in Section~\ref{sec:metric}, this may not be enough to protect privacy, since sensitive information that reveals identities can still be revealed by exploiting global distributions of the attributes \cite{li2007t}.
Here, we propose the Privacy-Preserving Attribute Selection (PPAS) Algorithm, a method to select and update facial attributes such that the distribution $S_E$ of any attribute $E$ is close to its real world distribution $S$ subject to constraint defined in Equation \ref{eqn:t-closeness}.
Unlike normal $t$-closeness, we further introduce a stochastic perturbation in the attribute selection process working toward $\epsilon$-differential privacy \cite{dwork2011differential}.
Our approach is formalized in Algorithm \ref{algo:PPAS} (for binary attributes).

\subsection*{Stage-III: Natural and Directed De-Identification}\label{sec:stage3}
To obfuscate facial images while preserve visual reality, we adopt a generative adversarial network (GAN) \cite{goodfellow2014generative}, which is designed as two players, $D$ and $G$, playing a minmax game with adversarial loss:
\begin{equation}
L_{adv} = \mathbb{E} [\log(D(\mathbf{x}))] + \mathbb{E} [\log(1 - D(G(\mathbf{x})))]
\end{equation}
where generator $G$ is trained to fool discriminator $D$, who tries to distinguish real images from adversarial ones.
GAN has been successful in many applications \cite{karras2018style,li2018semi} yet is notoriously difficult to train \cite{li2018toward}. As the face de-identification task can be categorize as a image-to-image translation problem, we customize the GAN model based on StarGAN \cite{choi2018stargan}, which has been widely demonstrated to have high usability along with realistic results, and the keys of which are attribute classification loss $L_{cls}$ and image reconstruction loss $L_{rec}$. $L_{adv}$, $L_{cls}$ and $L_{rec}$
forms the final objective function:
\begin{equation}
L = \lambda_1 L_{adv} + \lambda_2 L_{cls} + \lambda_3 L_{rec}
\end{equation}

\subsection*{Stage-IV: Adversarial Perturbation}\label{sec:stage4}
Adding a Gaussian noise to an image is generally considered as a simple and effective way to trick deep neural network-based detectors which have been showcased to be vulnerable against this attack.
There are also approaches that carefully craft perturbations for each data point. For example, \cite{goodfellow2014explaining} propose a fast gradient sign method to perturb an input through one step of gradient ascent. Furthermore, \cite{moosavi2017universal} shows there exists a universal adversarial perturbation that can cause images misclassified with high probability by state-of-the-art deep neural networks. The basic idea is formulated as follows: Suppose $\mu$ is the a distribution of images in $\Re^d$, $\hat{k}$ is a classifier that the output given an image $x$ is $\hat{k}(x)$. The universal perturbation vector $v \in \Re^d $ that can fool the classifier $\hat{k}$ should satisfy:
\begin{equation}
\|v\|_p \leq \xi
\end{equation}
\begin{equation}
\mathbb{P}_{x \sim \mu}(\hat{k}(x + v) \neq \hat{k}(x) \geq 1 - \delta)
\end{equation}
where $\xi$ limits the size of the universal perturbation vector, $\delta$ quantifies the failure rate of all the adversarial samples.

In this work, we introduce a universal perturbation vector as identified through an iterative approach. For each iteration $i$, we apply DeepFool \cite{moosavi2016deepfool} to identify the minimal perturbation to let $\hat{k}$ misclassify each input, and update the universal perturbation corresponding to hyperparameter $\epsilon_i$ to the total perturbation $v$.
It is shown that the algorithm works on a small portion of images sampled from the training dataset, and the universal perturbation generalizes well with respect to the data and the network architectures. Figure~\ref{fig:adversarial} outlines the proposed perturbation pipeline.
\input{fig_adversarial}

%% file: fig_stage1.tex
\begin{figure*}[h]
\begin{center}
\includegraphics[width=0.65\linewidth]{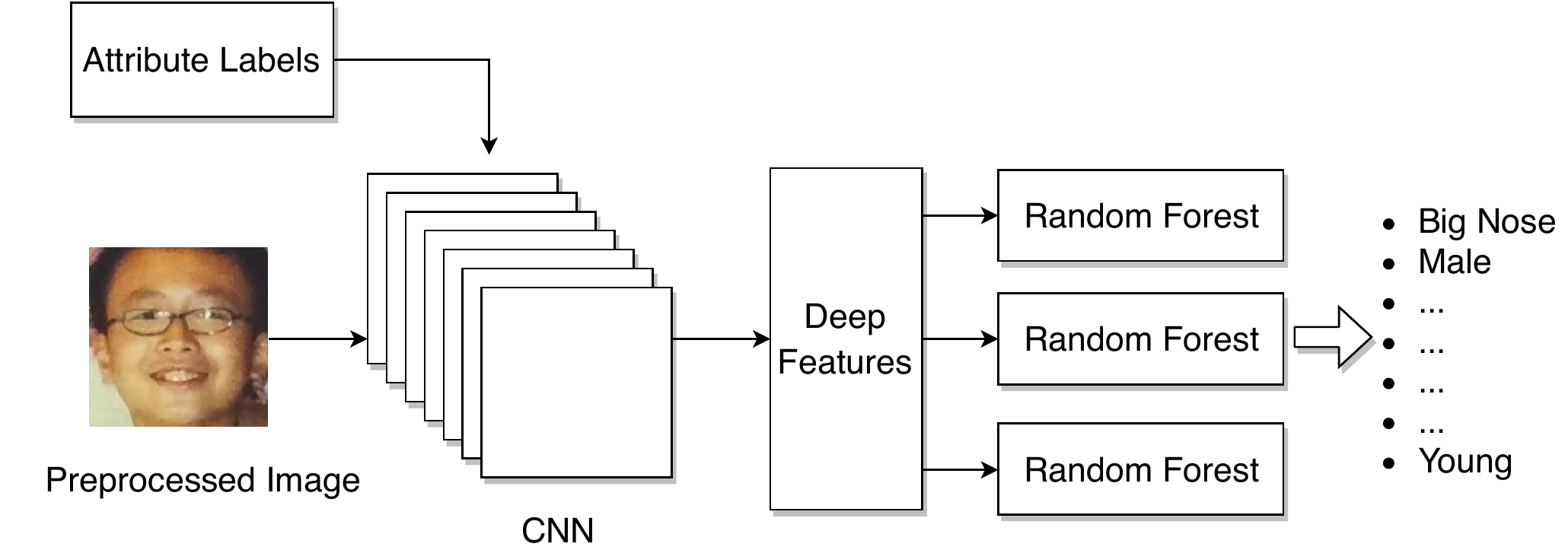}
\end{center}
\caption{Overview of the facial attribute prediction pipeline. We train a deep Convolutional Neural Network (CNN) fed by preprocessed and labelled images, and then extract deep features from the fully connected (FC) layer of the CNN and accordingly train random forest classifiers to predict facial attributes (the full attribute list can be found in Figure~\ref{fig:CeleA}).}
\label{fig:stage1}
\end{figure*}

%% file: algo_PPAS.tex
\begin{algorithm}[t]
\SetAlgoLined
\KwResult{Attribute set $\mathbb{A}''$.}
    Attribute set $\mathbb{A} \leftarrow \{ E_1, \dots, E_n \}$\;
    Attribute set $\mathbb{A}' \leftarrow \varnothing$ \;
    Size $N \leftarrow \mathbb{||A||}$ \;
    \For{$i = 1, \dots, N$}{
        \eIf{$d(S, S_{E_i}) \le t$}{
            Add attribute $E_i$ to $\mathbb{A}'$ \;
        }{
            Add attribute $\lnot E_i$ to $\mathbb{A}'$ \;
        }
    }
    \Return $\mathbb{A}'' \leftarrow Perturbation(\mathbb{A}', \epsilon)$ \;
\caption{The PPAS algorithm.}
\label{algo:PPAS}
\end{algorithm}

%% file: fig_adversarial.tex
\begin{figure}[H]
\begin{center}
\includegraphics[width=\linewidth]{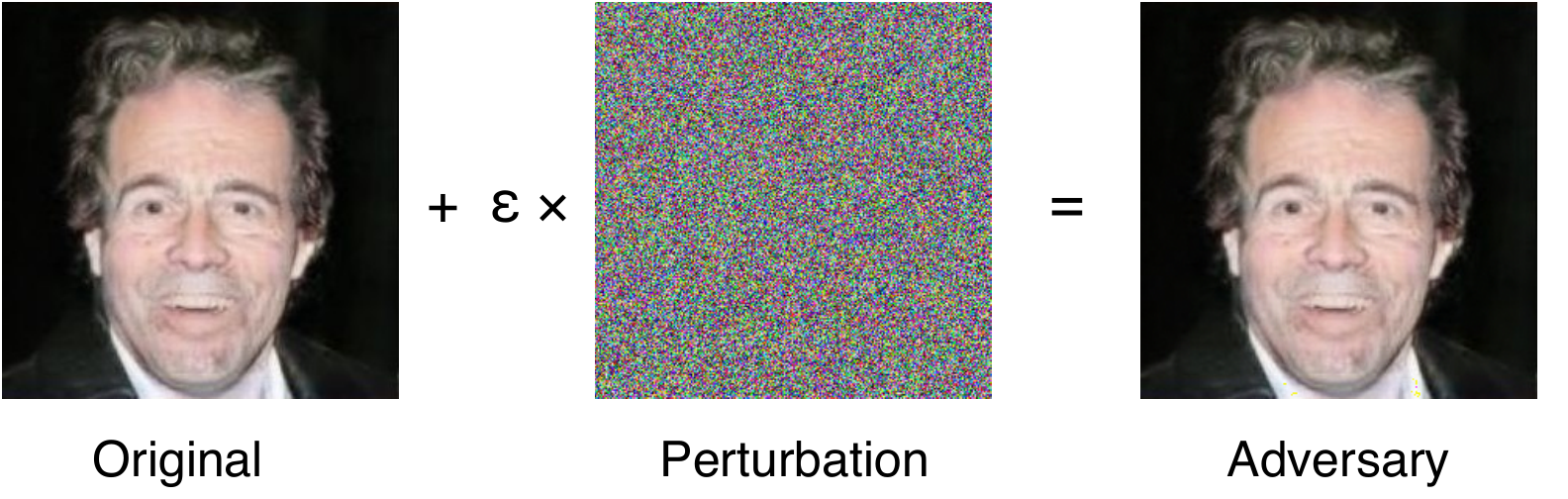}
\end{center}
\caption{An example of adversarial perturbation. In stage-IV, we introduce a small universal perturbation adjusted by parameter $\epsilon$ to synthesized images, tricking malicious detectors while preserving perceptual integrity.}
\label{fig:adversarial}
\end{figure}

%% file: sec_experiment.tex
\section{Experiment}\label{sec:experiment}

\input{fig_CeleA}
\subsection{Dataset}
We use Large-scale CelebFaces Attributes (CelebA) Dataset \cite{liu2015deep} for facial attribute estimation, which contains $202,599$ images and $10,177$ identities. Each image has $40$ attributes labels of boolean values, and their distributions have been shown in Figure~\ref{fig:CeleA}. Following the same protocol as described in \cite{liu2015deep}, we split the dataset into three folders: $160,000$ images ($8,000$ identities) for training; $20,000$ images ($1,000$ identities) for validation; and the rest $20,000$ images ($1,000$ identities) for testing.

\subsection{Image Preprocessing}
Before feeding the data into our deep models, we perform data preprocessing for each images in the datasets, including steps in order: face detection, landmark detection, alignment, and image cropping.
To obtain face landmarks, we deploy a Deep Alignment Network (DAN) \cite{kowalski2017deep}, which is a deep neural network of multiple stages and has demonstrated convincing performance even in extreme pose or lighting conditions in the wild.
Based on the $68$ landmarks provided by DAN for each image, we align and center the face in the image by calibrating positions of both left and right eyes.
Subsequently, we crop the images to a size of $256 \times 256$.
Figure \ref{fig:preprocessing} illustrates the preprocessing pipeline.
\input{fig_preprocessing}

\subsection{Training}
\paragraph{Attribute Estimation.}
As discussed in Section \ref{sec:stage1} earlier, a GoogLeNet \cite{szegedy2015going} is deployed for attribute estimation, using the same training settings as \cite{li2019beauty}. We adopt sigmoid cross-entropy as the loss function:
\begin{equation}
  L = - \frac{1}{n} \sum [ y \ln a + (1 - y) \ln (1 - a)]
\end{equation}
where $y$ is label and $a$ is output.
When training, we use a base learning rate of $10^{-5}$, which is reduced by a polynomial decay with a gamma of $0.5$.
Momentum is set to be $0.9$ and the weight decay is $2\times10^{-4}$.
$6\times10^{5}$ iterations with a batch size of $64$ are conducted for the training.
After extracting deep features from the FC layer, we train $40$ random forest classifiers for attribute estimation and achieve an accuracy that is comparable to current state-of-the-art \cite{rudd2016moon}.

\paragraph{Attribute Translation.}
After obtaining facial attributes that satisfies privacy constraints computed from previous steps, we employ StarGAN \cite{choi2018stargan} for face attributes translation and use CeleA \cite{liu2015deep} as the training set. We follow the settings in \cite{choi2018stargan}, where Wasserstein loss \cite{arjovsky2017wasserstein} is adopted to expedite the training process and the definition is as below:
\begin{align}
\mathcal{L}_{adv}
& = \mathbb{E}_x [D_{src}(X)] - \mathbb{E}_{x, c} [D_{src}(G(x, c))] \\
& = - \lambda_{gp} \mathbb{E}_{\hat{x}} [(|| \nabla_{\hat{x}} D(\hat{x}) ||_2 - 1)^2]
\end{align}
where $\hat{x}$ is uniformly sampled between a pair of original and synthesized images and we use $\lambda = 10$ here.
In terms of network architecture, we adopt a generator network of two convolutional layers for downsampling, six residual blocks \cite{he2016deep}, and two transposed convolutional layers for upsampling; and we use PatchGANs \cite{isola2017image} as the discriminator network for binary classification (i.e., whether certain image patches are fake or real).

\input{fig_evaluation}
\subsection{Evaluation}
Figure~\ref{fig:intro} compares our approach with canonical face obfuscation methods, showcasing a significant improvement in photo-reality and a success in face de-identification.
Figure~\ref{fig:evaluation} illustrates experimental results in pairs, where left is the original image and right is the result generated by our framework.
It demonstrates that face identities are preserved in a perceptually natural manner, and in the meantime, each pair of images still shares certain common attributes in correspondence with various privacy policies and application scenarios (for example, pair $(1, 4)$ differs in hair color, lip color, gender, and age, but shares the same eye sizes; pair $(1,1)$ only differs in gender and leaves other semantic attributes identical). Also, artifacts introduced by the proposed the adversarial perturbation are shown to be visually negligible, even though they have been upscaled for illustrative purposes.
Table \ref{tab:metric} compare image quality under widely used perceptual metrics. Being aware that these model-based metrics fail to capture many nuances of human perception \cite{zhang2018unreasonable} and a smaller or larger value does not necessarily imply higher or lower image quality \cite{li2017single}, we list the results here only for reference purpose.
\input{tab_metric}

%% file: fig_CeleA.tex
\begin{figure*}[t]
\begin{center}
\includegraphics[width=\linewidth]{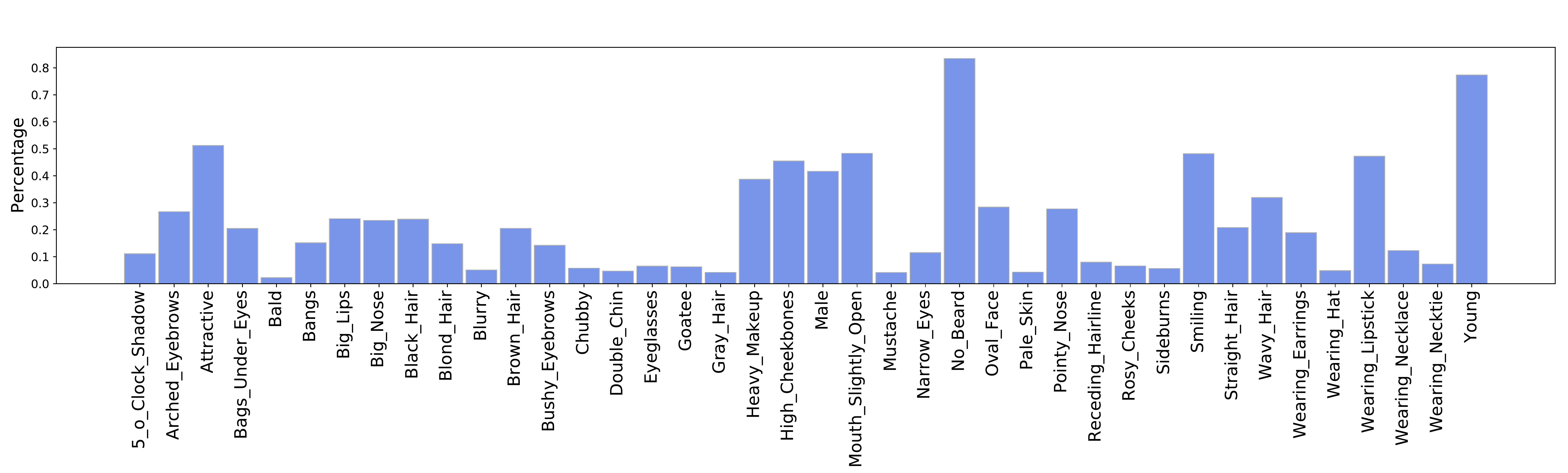}
\end{center}
\caption{Facial attributes and their distributions in the CeleA dataset \cite{liu2015deep}.}
\label{fig:CeleA}
\end{figure*}

%% file: fig_preprocessing.tex
\begin{figure}[H]
\begin{center}
\includegraphics[width=1.05\linewidth]{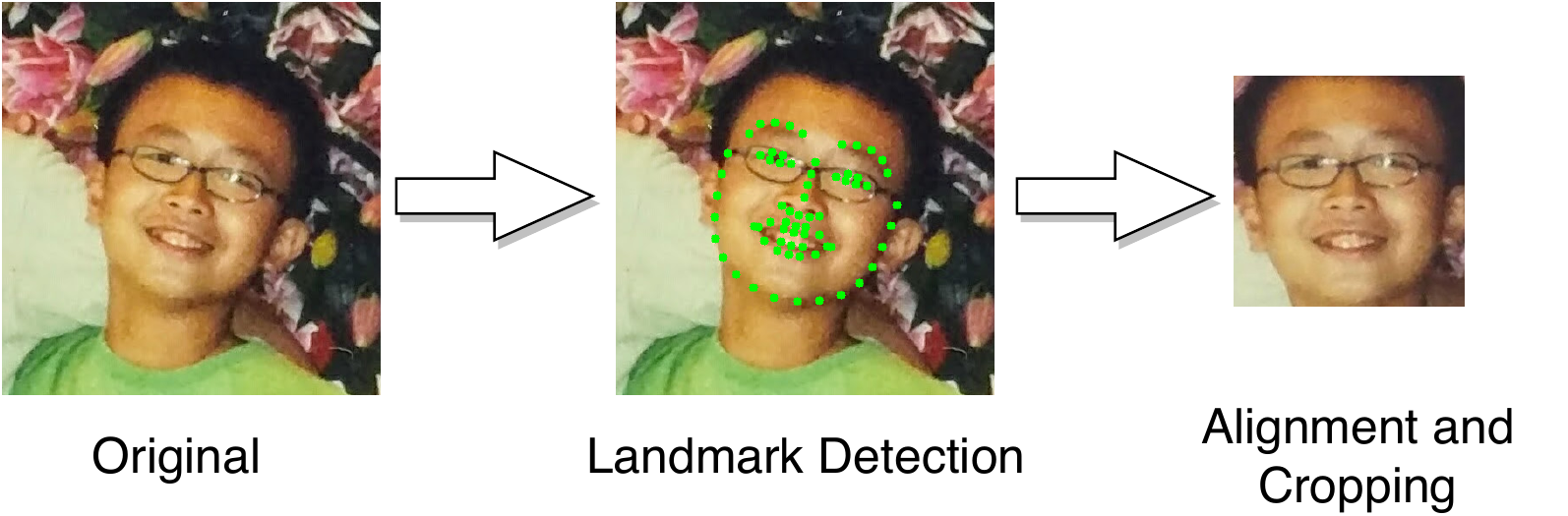}
\end{center}
\caption{Image preprocessing pipeline. We deploy a Deep Alignment Network (DAN) \cite{kowalski2017deep} to obtain facial landmarks, based on which we accordingly align faces and crop images.}
\label{fig:preprocessing}
\end{figure}

%% file: fig_evaluation.tex
\begin{figure*}[!t]
\begin{center}
\includegraphics[width=0.1165\textwidth]{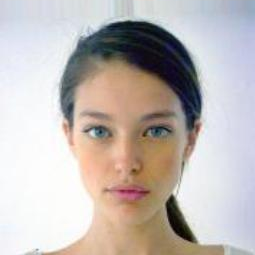}
\includegraphics[width=0.1165\textwidth]{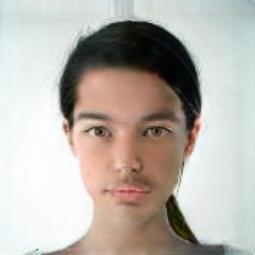}
~
\includegraphics[width=0.1165\textwidth]{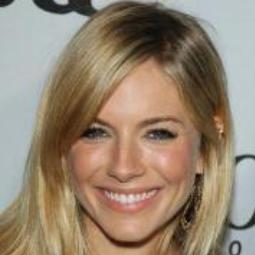}
\includegraphics[width=0.1165\textwidth]{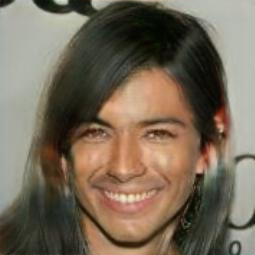}
~
\includegraphics[width=0.1165\textwidth]{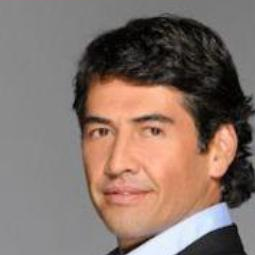}
\includegraphics[width=0.1165\textwidth]{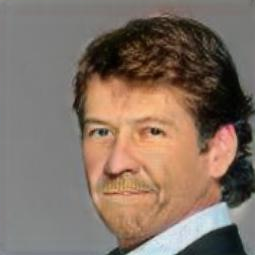}
~
\includegraphics[width=0.1165\textwidth]{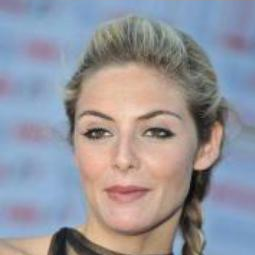}
\includegraphics[width=0.1165\textwidth]{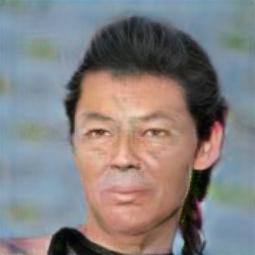}
\\
\includegraphics[width=0.1165\textwidth]{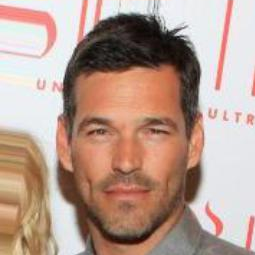}
\includegraphics[width=0.1165\textwidth]{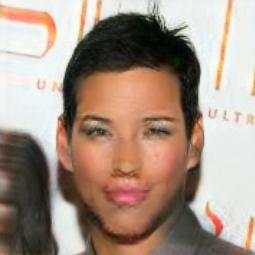}
~
\includegraphics[width=0.1165\textwidth]{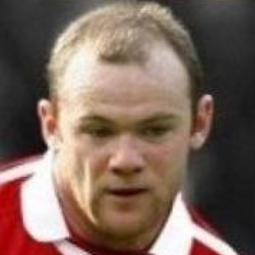}
\includegraphics[width=0.1165\textwidth]{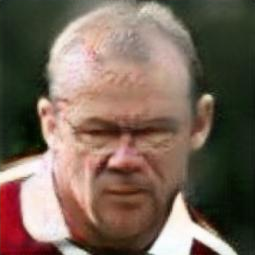}
~
\includegraphics[width=0.1165\textwidth]{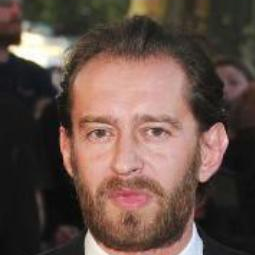}
\includegraphics[width=0.1165\textwidth]{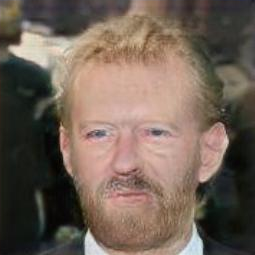}
~
\includegraphics[width=0.1165\textwidth]{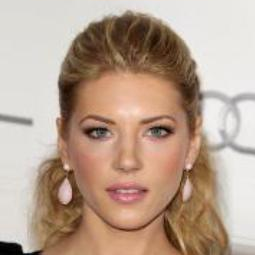}
\includegraphics[width=0.1165\textwidth]{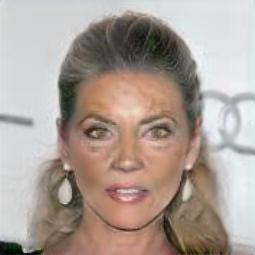}
\\
\includegraphics[width=0.1165\textwidth]{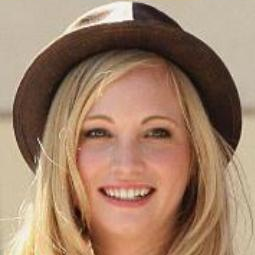}
\includegraphics[width=0.1165\textwidth]{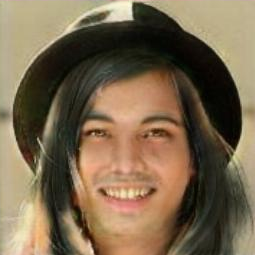}
~
\includegraphics[width=0.1165\textwidth]{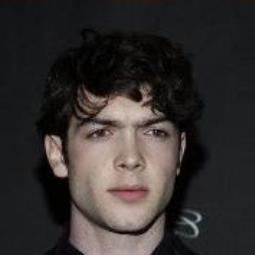}
\includegraphics[width=0.1165\textwidth]{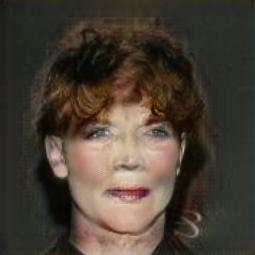}
~
\includegraphics[width=0.1165\textwidth]{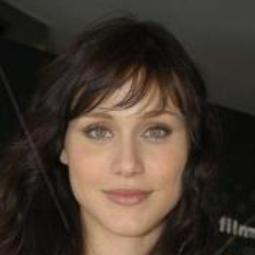}
\includegraphics[width=0.1165\textwidth]{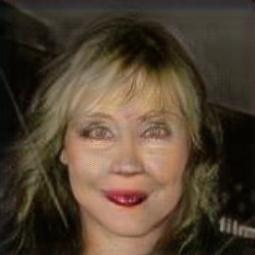}
~
\includegraphics[width=0.1165\textwidth]{img/npc_67_11_1.png}
\includegraphics[width=0.1165\textwidth]{img/npc_67_11_5.png}
\\
\includegraphics[width=0.1165\textwidth]{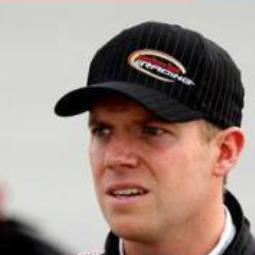}
\includegraphics[width=0.1165\textwidth]{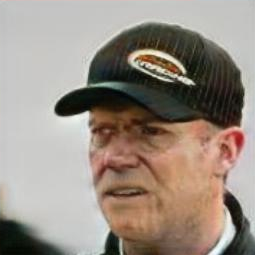}
~
\includegraphics[width=0.1165\textwidth]{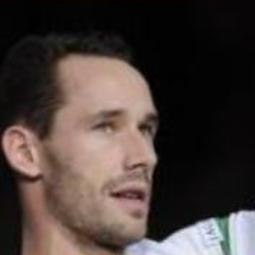}
\includegraphics[width=0.1165\textwidth]{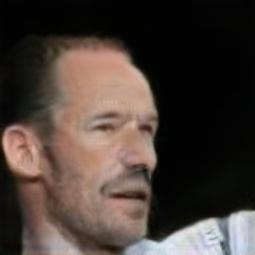}
~
\includegraphics[width=0.1165\textwidth]{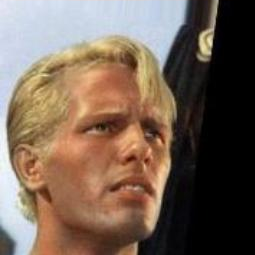}
\includegraphics[width=0.1165\textwidth]{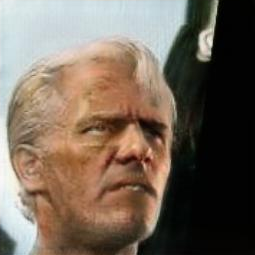}
~
\includegraphics[width=0.1165\textwidth]{img/npc_105_11_5.png}
\includegraphics[width=0.1165\textwidth]{img/npc_105_11_6.png}
\end{center}
\caption{Some experimental results. In each pair, left is the original image and right is the synthesized result with an altered identity. The results show that face identities are preserved in a perceptually natural manner, and in the meantime, each pair of images still shares certain common attributes in correspondence with various privacy policies and application scenarios. Furthermore, artifacts introduced by the proposed the adversarial perturbation are shown to be visually negligible (the perturbation has been intentionally upscaled here for illustrative purposes).}
\label{fig:evaluation}
\end{figure*}

%% file: tab_metric.tex
\begin{table}[H]
\begin{center}
\resizebox{\columnwidth}{!}{
\begin{tabular}{cccccc}
\hline
    & Blurring  & Pixelation & Masking & Inpainting & Ours \\
\hline
PSNR \cite{huynh2008scope}  & 24.532 & 22.802 & 15.459 & 18.097 & 20.079\\
SSIM \cite{wang2004image} & 0.8281 & 0.8226 & 0.8762 & 0.8020 & 0.7894 \\
MS-SSIM \cite{wang2003multiscale} & 0.8842 & 0.8784 & 0.8204 & 0.7659 & 0.8650 \\
\hline
\end{tabular}
}
\end{center}
\caption{Image quality comparison under different metrics.}
\label{tab:metric}
\end{table}

%% file: sec_conclusion.tex
\section{Conclusion and Future Work}\label{sec:conclusion}
To naturally obfuscate face identities while preserves privacy in a controllable and measurable manner, we proposed the AnonymousNet framework, which consists of four stages:
facial feature extraction, semantic-based attribute obfuscation, de-identified face generation, and adversarial perturbation.
This framework successfully overcomes the shortages of existing methods - being able to generate photo-realistic images with fake identity and capable of balancing privacy and usability both qualitatively and quantitatively. It also answers questions such as ``is it private now?'', ``how private is it?'', and ``can it be more/less private?'' counterfactually.
Considering the threats from adversaries, especially the malicious detectors that are prevalent in today's Internet, we further introduced a universal adversarial perturbation so as to trick other deep neural networks as much as possible.
Experimental results support the effectiveness of our approach by showing photo-realistic results with negligible artifacts.

In the future, we would like to evaluate perturbation performance among different deep neural network-based detectors qualitatively and quantitatively, which is ignored here due limitations in space and computational resources.
We are also interested in robustness, scalability, and extensibility of this framework under various real-world settings.
In the experiments, we find that different facial attributes vary in ``distinguish power'', i.e., attributes such as \texttt{Age} and \texttt{Gender} are perceptually more powerful in helping distinguish an identity than \texttt{Cheekbones\_Height}, which align with our intuitions.
This advises a future research direction that a user study can be made to explore these differences qualitatively and quantitatively, and in the end, figure out ``the crux of facial indistinguishability''.

%% file: sec_acknowledgment.tex
\section*{Acknowledgment}
The authors especially thank Professor Chris Clifton for insightful discussions in differential privacy and privacy metrics in the context of facial images.
We are also grateful for Professor Fengqing Maggie Zhu and anonymous reviewers from the CV-COPS'19 program committee for their comments and helpful suggestions.
The idea of this paper was born when the first author writing \cite{liu2019understanding} during his research internship at ObEN, Inc. He thanks colleagues at ObEN for the wonderful times they shared.